\documentclass[journal,comsoc]{IEEEtran}
\usepackage[T1]{fontenc}
\usepackage{multicol}  
\usepackage{multirow}  
\usepackage{tabu}  
\usepackage{booktabs}
\usepackage{threeparttable}
\usepackage{latexsym}
\usepackage{setspace}
\usepackage{footnote}
\usepackage{tablefootnote}
\usepackage{graphicx}
\usepackage{color}
\graphicspath{{./Figures/}}

\ifCLASSINFOpdf
  \graphicspath{{./Figures/}}
\else
\fi
\usepackage{amsmath}
\usepackage[cmintegrals]{newtxmath}
\begin{document}
%
\title{A Unified Structure for Efficient RGB and RGB-D Salient Object Detection}
%
%
%

\author{Peng~Peng,
	and~Yong-Jie~Li,~\IEEEmembership{Senior Member,~IEEE}
\thanks{P. Peng, Yong-Jie. Li are with the MOE Key Laboratory for Neuroinformation, the School of Life Science and Tecnology, University of Electronic Science and Technology of China, Chengdu 610054, China.
	E-mail: pengpanda.uestc@gmail.com, liyj@uestc.edu.cn}
\thanks{Corresponding author: Yong-jie Li.}
\thanks{Manuscript received April 19, 20XX; revised August 26, 20XX.}}

%
%

\markboth{Journal of \LaTeX\ Class Files,~Vol.~14, No.~8, August~20XX}%
{Shell \MakeLowercase{\textit{et al.}}: Bare Demo of IEEEtran.cls for IEEE Communications Society Journals}
%



\maketitle

\begin{abstract}
	Salient object detection (SOD) has been well studied in recent years, especially using deep neural networks. 
However, SOD with RGB and RGB-D images is usually treated as two different tasks with different network structures that need to be designed specifically.
In this paper, we proposed a unified and efficient structure with a cross-attention context extraction (CRACE) module to address both tasks of SOD efficiently.
The proposed CRACE module receives and appropriately fuses two (for RGB SOD) or three (for RGB-D SOD) inputs.
The simple unified feature pyramid network (FPN)-like structure with CRACE modules conveys and refines the results under the multi-level supervisions of saliency and boundaries.
The proposed structure is simple yet effective; the rich context information of RGB and depth can be appropriately extracted and fused by the proposed structure efficiently.
Experimental results show that our method outperforms other state-of-the-art methods in both RGB and RGB-D SOD tasks on various datasets and in terms of most metrics. 

\end{abstract}


%
\IEEEpeerreviewmaketitle

\IEEEPARstart{H}uman visual system can focus on a target object with little movement in a scene. 
This ability helps humans search and find targets efficiently.
Inspired by this fact, researchers proposed a family of methods \cite{achanta2009frequency, ChengPAMI,shi2016hierarchical,yang2016unified,peng2016salient} using handcrafted features to accomplish the task of salient object detection (SOD). 
SOD is believed to facilitate image segmentation \cite{rahtu2010segmenting}, tracking \cite{Ma2017A}, retrieval \cite{gao2015database}, compression \cite{ji2013video} and other computer vision tasks.

Deep learning has dominated the field of SOD since the use of the convolutional neural network (CNN) in pioneering works \cite{li2015visual, zhao2015saliency}.
The fully convolutional network (FCN) has addressed the problem of pixel-level classification of images with the utilization of an upsampling layer in the decoder stage.
The FCN has shown advantages in many research fields, especially image segmentation \cite{long2015fully}.
Therefore, SOD has also been remarkably facilitated after the introduction of the FCN \cite{wang2016saliency}. 

However, suffering from pooling--upsampling layers, many image details are lost, which means that the results are usually fuzzy and incomplete.  
The feature pyramid network (FPN) \cite{lin2017feature} structure was proposed to solve this problem by adding connections between the encoder and decoder, and has become popular in image segmentation tasks.
Many methods \cite{qin2019basnet, wang2019inferring} for SOD have adopted FPN-like structures and achieved good performance.
However, due to the overly simple structure of the decoder and the connections, segmentation problems still exist, and obtaining clear and accurate results remains challenging.

As aforementioned, an attention mechanism is a natural component of the human visual system.
Interestingly, artificial neural networks also seem to have such a mechanism when processing natural language and images \cite{vaswani2017attention}.
The utilization of attention in the DNN has led to breakthroughs in many research fields, including improved SOD performance.
However, simply applying spatial attention or channel attention is not enough to find the salient object; meanwhile, the non-local (attention) \cite{wang2018non} costs too much in both time and space.

SOD in RGB-D images is usually treated as a task different from RGB image SOD, and most of the existing methods are specifically designed for RGB-D SOD \cite{cong2018review,zhou2020rgb}. 
However, RGB-D SOD is just an extension of RGB SOD with an additional modal, and thus there should be a structure or framework to unify these two tasks.

\begin{figure}
	\begin{center}
		\includegraphics[width=0.45\textwidth]{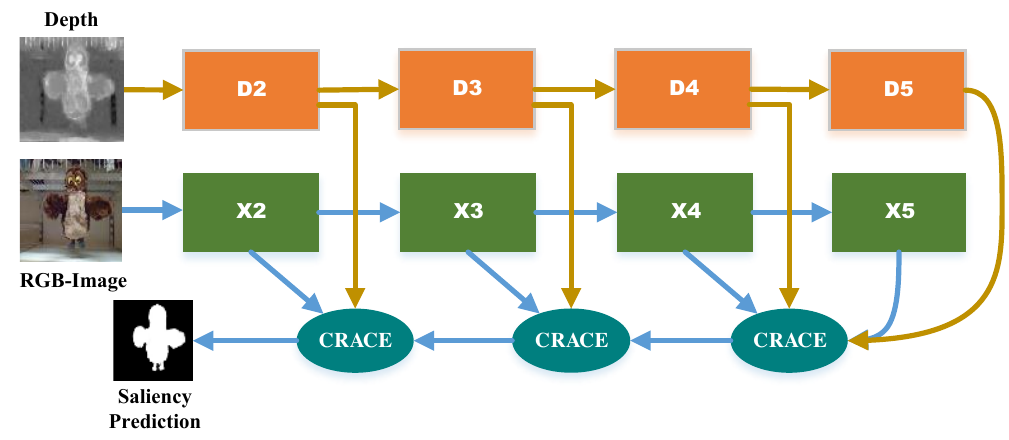}
	\end{center}
	\caption{Overview of the proposed structure for RGB image (with the blue arrows) and RGB-D image SOD (with all blue and orange arrows). The proposed cross-attention context extraction (CRACE) module can extract and fuse cross-level features and even cross-modal features. }
	\label{fig_unified}
\end{figure}

In this work, to overcome the aforementioned problems, we proposed a cross-attention context extraction (CRACE) module to serve as the decoder of an FPN-like network, as shown in Fig.\ref{fig_unified}.
This structure is designed as a unified framework to accomplish both tasks of RGB image SOD and RGB-D image SOD.
When dealing with RGB-D features, depth information is simply extracted with a backbone network (usually VGG \cite{Simonyan15} or ResNet \cite{he2016deep}) and then fed into the CRACE module.

In this CRACE module, we first design a cross-attention block to extract the attention map from high-level global features and low-level local features, and then optimize the features so that we can use the attention map.
This block is also designed to extract and fuse cross-modality features.
Except for cross-level RGB features, depth features can also be extracted and fused in this block.
It may also be possible to extract and fuse other modalities, such as optical flow, but this is beyond the scope of this work.

\begin{figure*}
	\begin{center}
		\includegraphics[width=1\textwidth]{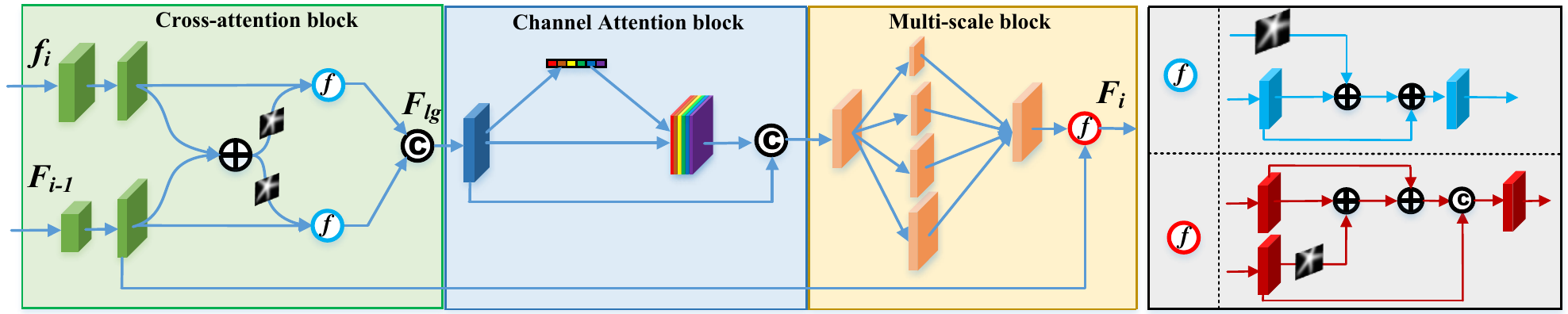}
	\end{center}
	\caption{Cross-attention context extraction (CRACE) module. The feature blocks of $ f_i $ ($ f_l $) and $ F_{i-1} $ ($ f_g $) (and depth feature $ d_i $) are first fused in the cross-attention block and then optimized in the channel attention block; more information is extracted in the multi-scale block and attentive fusion block. The output of the CRACE module ($ F_i $) then serves as one of the inputs of the next-level CRACE module.}
	\label{fig:CCE}
\end{figure*}

After feature optimization, a channel attention block, a multi-scale operation, and attentive fusion are implemented to extract rich context information.
The overall structure of the CRACE module is shown in Fig. \ref{fig:CCE}.
These context features are conveyed and fused from global-level deep layers to local-level shallow layers in the proposed network like a kind of information flow.
The source code and results will be available on our website$\footnote{http://www.neuro.uestc.edu.cn/vccl/publications.html}$.
In summary, our main contributions are as follows:

1) We proposed a cross-attention context extraction module  that appropriately extracts and fuses the coarser global feature and the finer local features, including cross modality features such as depth information.

2) Based on this module, we proposed a simple unified FPN-like structure for RGB and RGB-D SOD; this structure conveys and fuses the context information from global to local, or cross-modality features under the supervision of ground-truth and object boundary.

3) Quantitative experiments on five popular RGB image SOD datasets and seven RGB-D image SOD datasets (and using seven metrics) demonstrate that our model achieves state-of-the-art (SOTA) performance, thereby validating the effectiveness of our model on both tasks. 

\section{Related work}
Since the pioneering work of \cite{Itti1998A}, there has been tremendous progress in SOD.
There are many methods based on some cognitive hypothesis to process extracted handcrafted features for SOD \cite{achanta2009frequency,ChengPAMI,shi2016hierarchical}.
Some researchers adopted informatics or mathematics while others used machine learning to segment salient objects \cite{peng2016salient,Yuan2017Reversion}.
Overviews of the use of these handcrafted methods in RGB image SOD can be found in \cite{borji2014salient, cong2018review}, and the handcrafted methods employed in RGB-D image SOD can be found in \cite{peng2014rgbd, zhou2020rgb}.
Since there is a great number of methods based on deep learning in this field, we only review some of the deep learning based methods in terms of RGB and RGB-D image SOD.
\paragraph{RGB Salient Object Detection}
\cite{zhao2015saliency} is one of the earliest works treating saliency detection as a high-level task and  exploiting a pre-trained deep model to capture the local and global contexts of images.
Hou \textit{et al.} \cite{hou2017deeply} adopted short connections to integrate high-level location knowledge and low-level structure information.
Liu \textit{et al.}\cite{liu2018picanet} proposed a pixel-wise contextual attention and then integrated it into a U-Net to detect saliency.
Meanwhile, the recurrent fully convolution network was leveraged in \cite{wang2018salient} to automatically learn and refine the results, and reversed saliency was adopted to refine the finer structure of salient objects in \cite{chen2020reverse}.

Information including fixation, semantics, and edges has also been adopted in many methods.
Wang \textit{et al.} \cite{wang2016saliency} jointly trained their model for fixation and SOD in a convolutional LSTM network and obtained good performance.
Semantic prior \cite{nguyen2019semantic} and captioning \cite{zhang2019capsal} were leveraged to detect salient objects from cluttered scenes.
Boundaries are frequently used in many SOD methods \cite{ li2018contour,liu2019simple,qin2019basnet} and usually help produce better results.
More reviews of RGB image SOD in the deep learning era can be found in \cite{borji2019saliency,cong2018review,wang2019salient}.

\paragraph{RGB-D Salient Object Detection}
Depth information is important in many tasks and plays a role in RGB-D SOD.
\cite{qu2017rgbd} fed their deep model handcrafted features and produced more accurate results than traditional methods.
Piao \textit{et al.} \cite{piao2019depth} proposed a depth-induced multi-scale network to fully extract the complementary cues from RGB and depth information.
Meanwhile, Zhao \textit{et al.} \cite{zhao2019contrast} introduced the contrast prior into a deep model to enhance the depth information, and then integrated it with RGB information.
Piao \textit{et al.} \cite{Piao2020A2deleAA} proposed a depth distiller to transfer the depth knowledge to RGB features. Furthermore, Fan \textit{et al.} \cite{Fan2020RethinkingRS} proposed an efficient three-stream structure and depth depurator to learn cross-modal features.
Selective self-mutual attention was learned and fused in a two-stream CNN model in \cite{Liu2020LearningSS}.
Inspired by the data labeling process, Zhao \textit{et al.} \cite{Zhang2020UCNetUI} proposed a conditional variational auto-encoder to model annotation uncertainty and produce saliency maps.
More reviews of RGB-D image SOD can be found in \cite{cong2018review,zhou2020rgb} 

All of the methods mentioned above are designed for a single task of either RGB or RGB-D image SOD.
There are very few deep learning-based methods designed for both tasks using the same structure.
Zhao \textit{et al.} \cite{Zhao2020SuppressAB} proposed a gated dual branch structure to transfer valuable information from the encoder to the decoder, and produced accurate RGB and RGB-D saliency maps.
Zhao \textit{et al.} \cite{Zhao2020IsDR} proposed a model where the depth maps were utilized in the training process but not in the test process, which allowed the model to compute both RGB and RGB-D saliency maps.

However, these methods are either too complicated in regards to the structure and training process or achieve suboptimal accuracy.
Our method adopts a basic FPN structure with three proposed CRACE modules and achieves competitive results with SOTA methods.

\section{The Proposed Network}
In this section, we proposed a unified network based on a cross-attention context extraction (CRACE) module for RGB and RGB-D image SOD.

\subsection{Cross-attention Context Extraction Module}
To extract the context information from the features derived by the backbone network, we proposed a CRACE module.
The CRACE module consists of three main blocks and one fusion operation, i.e., a cross-attention block, a channel attention block, a multi-scale block, and attentive fusion.

\begin{figure*}
	\begin{center}
		\includegraphics[width=0.9\textwidth]{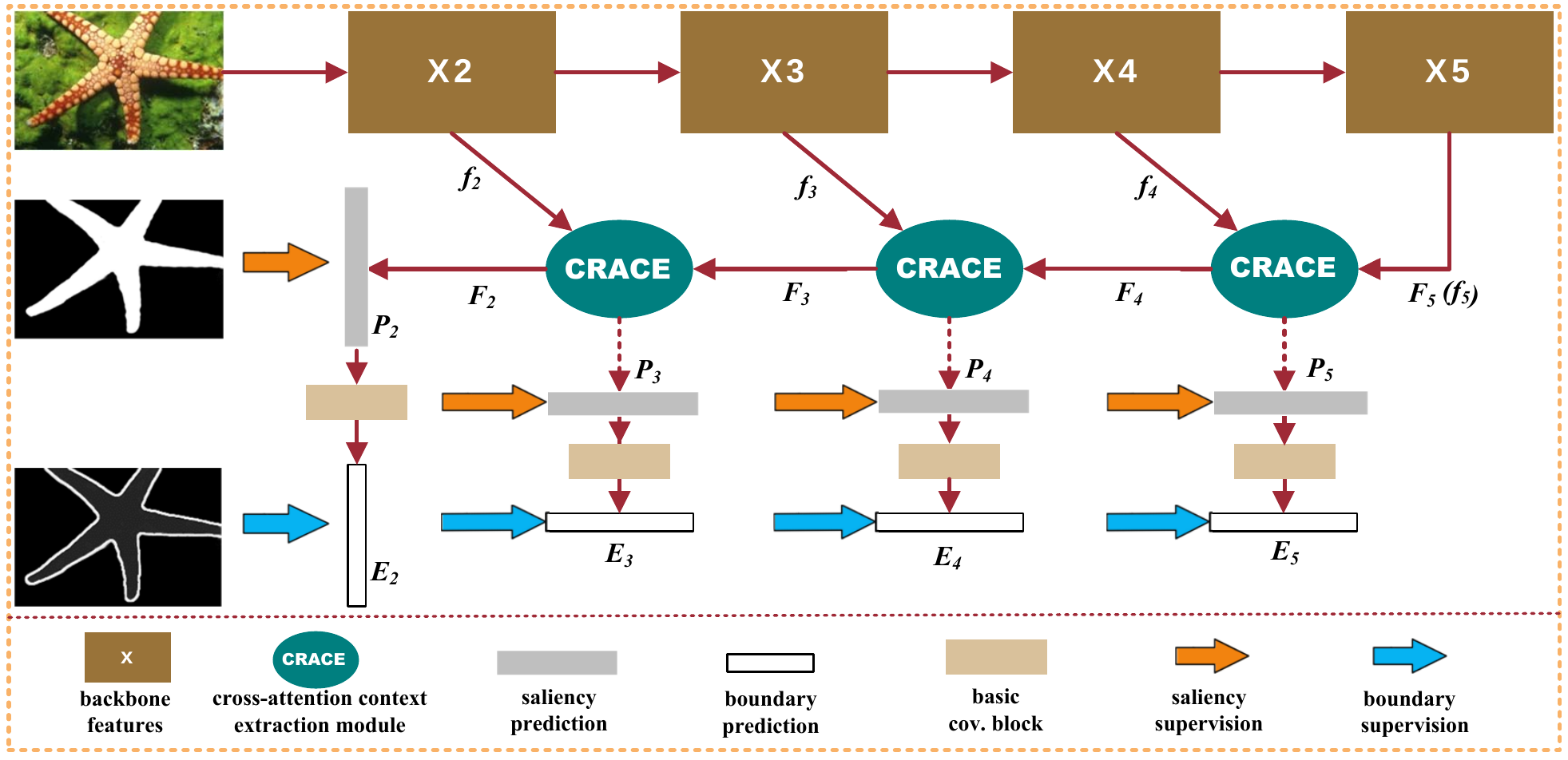}
	\end{center}
	\caption{The architecture of the proposed unified structure dealing with the task of RGB image SOD (the part indicated by blue arrows in Fig.\ref{fig_unified}). Multi-level saliency and boundary supervisions are also adopted in the training stage.  For the task of RGB-D image SOD, the depth map is processed by the backbone network and then the multi-level features are fed into the CRACE modules following the orange arrows in Fig.\ref{fig_unified}. Please see the text for more details.}
	\label{fig:CFNet}
\end{figure*}

\paragraph{Cross-attention Block} 
For a regular neural network the features extracted in the deep layers contain more \textit{global-level} semantic information of images while the shallow layers contain more \textit{local-level} structure information (such boundary information) \cite{Ronneberger2015UNetCN,lin2017feature}.
The widely used spatial attention block extracts the attention map in a single layer, which usually results in either a loss of structural information in the deep layers or a loss of semantic information in the shallow layers.

The proposed cross-attention block addresses this problem by extracting and fusing two different layers.
In this block, two feature blocks (usually one shallower $ f_l $ and one deeper $f_g$ layer) are first converted to the same channel by some basic conventional blocks. Then, we add them up and reduce the number of channels to 1 to produce a cross-attention map $ A_{lg} $.
This cross-attention map contains attention information for both feature layers: more fine structural information for the deeper layer and more semantic information for the shallower layer.
This mechanism can be formulated as
\begin{equation}\label{Eq.CCE_CA} A_{lg} =\sigma (C^{n\rightarrow 1}_{1\times 1} [C_l(f_l)+C_g(up(f_g))])  \end{equation}
\begin{equation}\label{Eq.CCE_FL} f'_l = C_l(f_l) + A_{lg}\cdot C_l(f_l) \end{equation}
\begin{equation}\label{Eq.CCE_FG} f'_g =  C_g(up(f_g)) + A_{lg}\cdot C_g(up(f_g)) \end{equation}
\begin{equation}\label{Eq.CCE_out} F_{lg} =  concate(f'_l, f'_g)  \end{equation}

where these $ C(\cdot) $ are conventional blocks combined with batch normalization and ReLU, $ \sigma $ is the Sigmoid function, and $ C^{n\rightarrow 1}_{1\times 1} $ is the convolution operation with a filter size of 1$ \times $1. We reduce the number of channels from $ n $ to 1, $ up(\cdot) $ is the up-sampling operation, and $ n $ is set to 64 in our model (taking into consideration accuracy and efficiency).  

Our CRACE module is different from the cross feature module (CFM) proposed in \cite{wei2019f3net}.
We extract both global-level semantic and local-level structural information to create a cross-attention mechanism to guide the optimization of different-level features; in contrast, CFM simply multiplies and adds up the features.  
Furthermore, our CRACE module is designed to expand so that it can process multiple modalities when additional input is added, such as depth information.

\paragraph{Channel Attention Block}
Channel attention is widely used in many methods as it is simple but effective \cite{Fan2020RethinkingRS,Zhao2020SuppressAB} in selecting features by their weightings.
Different from CBAM \cite{woo2018cbam}, we combine a channel attention module after the cross-attention block to select the optimal output of features.
Unlike the conventional channel attention block, we adopt a residual form to preserve more information and compute the feature maps $ F_{ch} $, which can be formulated as 
\begin{equation}\label{Eq.CCE_CH} F_{ch} =C^{2n\rightarrow n}_{1\times 1} (concate[GAP(F_{lg})\cdot F_{lg}, F_{lg}]) \end{equation}
where $ GAP(\cdot) $ is the channel-wise global average pooling, and $ C^{2n\rightarrow n}_{1\times 1} $ is the operation to reduce the number of channels to $ n $.
This block selects and assigns the weights of each channel for the features, which facilitates context information extraction.

\paragraph{Multi-scale Block}
The usefulness of the multi-scale strategy has been validated in many methods \cite{chen2020reverse, Zhao2020SuppressAB}; it can extract more information from multi-scale features.
To exploit this ability, we apply a multi-scale operation before the attentive fusion block. 
The structure of this block is shown in Fig. \ref{fig:CCE} and formulated as 
\begin{equation}\label{Eq.CCE_Fms} F_{ms} = \sum_{j=1}^{4}Branch_j(F_{ch})\end{equation}
where the $ Branch_j,  j=\{1,2,3,4\} $ are the operations to extract multi-scale features; these operations are down-sampling, convolution, up-sampling, and finally their combination.
In the down- and up-sampling operations, the sampling rates are set to $ \{1, 2, 4, 8\} $, and the convolution filter sizes are 3 with dilation rates $\{1,4,6\}  $ to strike a balance between accuracy and efficiency. 

\paragraph{Attentive Fusion}
Finally, the outputs of the multi-scale block are attentively fused using the global layer feature $ f_g $.
The fused output $ F $ is designed as
\begin{equation}\label{Eq.CCE_F} F = C_{1\times 1}^{2n\rightarrow n }(concate(A_g\cdot f_g+f_g, F_{ms}))\end{equation}
where $ C_{1\times 1}^{2n\rightarrow n } $ is a convolution operation with a filter size of $ 1\times 1 $ to reduce the number of channels, and $ A_g $ is the spatial attention map of $ f_g $, which is calculated as
\begin{equation}\label{Eq.CCE_Ag} A_g = \sigma (C^{n\rightarrow 1}_{1\times 1} (f_g))\end{equation}
$ \sigma $ and $  C^{n\rightarrow 1}_{1\times 1}(\cdot)$ are the same as in Eqn.\ref{Eq.CCE_CA}.
This fusion operation leverages higher global information and fuses this information into the local features.

\subsection{Network Architecture}
We adopt the widely used backbone network ResNet50 \cite{he2016deep} as the feature extractor. We pre-trained the backbone on ImageNet \cite{deng2009imagenet}, as in most of the literature.

\paragraph{Context Flow}
Our network is compact, with a simple connection of the three CRACE modules in an FPN-like structure. This is illustrated in Fig.\ref{fig:CFNet}.
Only the output of the last four encoders are processed, as in many other methods \cite{chen2020reverse,wei2019f3net}, since the first block is computationally expensive.
Therefore, the output features ($f_2\sim f_5  $) of each encoder are the inputs of the proposed CRACE modules.

The features of $ f_4 $ and $ f_5 $ are first input into the CRACE module to extract the fused feature $ F_4 $ that contains the global context information and local structural features.
Then $ F_4 $ serves as the global input feature of the next CRACE module.
Consequently, context information flows from deep to shallow.
If we treat the operation of the CRACE module as a function $ CRACE(\cdot) $, it can be formulated as follows:
\begin{equation}\label{Eq.decoder} F_{i-1} =  CRACE(f_{i-1}, F_i)  \end{equation}
where $ i\in \{3,4,5\} $, and $ F_5 $ is the result of $ f_5 $ after a convolution operation, which occurs inside the CRACE block.

\subsection{Loss Function}
Jointly training boundaries and multi-level supervision is common in SOD methods, and this usually allows us to obtain better performance compared to when we employ single-supervision methods \cite{wei2019f3net,liu2019simple}.

\paragraph{Multi-level supervision}
Following the idea of multi-level supervision, we train the proposed network by adopting multi-level supervision with ground-truth $ S $ using  
\begin{equation}\label{Eq.S_spvs} L_S  = \sum_{i=2}^{5}l_s(P_i, S)   \end{equation}
where the multi-level saliency predictions $ P_{i} $ of the model are calculated from the features as demonstrated in Eqn.\ref{Eq.decoder} using a basic convolution block:
\begin{equation}\label{Eq.Prediction} P_{i} =  Conv(F_{i})  \end{equation}

The loss function $ l_s $ in Eqn.\ref{Eq.S_spvs} is the combination of conventional binary cross entropy (BCE) loss and IoU loss \cite{wei2019f3net,chen2020reverse,Zhao2020IsDR}.
Let $ P, S \in \mathbb{R}^{W\times H\times 1}$ be the saliency prediction from any level of the CRACE module and ground-truth maps, and the BCE loss and IoU loss are defined as 
\begin{equation}\label{Eq.L_bce} L_{bce} = -\sum_{j=1}^{H}\sum_{k=1}^{W}P_{jk}log(S_{jk})  \end{equation}
\begin{equation}\label{Eq.L_iou} L_{iou} = 1-\frac{\sum_{j=1}^{H}\sum_{k=1}^{W}P_{jk}\cdot S_{jk} +1}{\sum_{j=1}^{H}\sum_{k=1}^{W}(P_{jk}+S_{jk}-P_{jk}\cdot S_{jk})+1}   \end{equation}

We also jointly train the proposed network with the boundaries $ E $ generated from the ground-truth maps by 
\begin{equation}\label{Eq.Edge} E  = S - erode(S)   \end{equation}
where $ erode(\cdot) $ is the image erosion operation \cite{zhou2020interactive}.
The supervision of boundaries can be formulated as
\begin{equation}\label{Eq.E_spvs} L_E  =  \sum_{i=2}^{5}l_e(E_i, E)   \end{equation}
where $ E_i $ is the boundary features of $ F_i $ after a basic convolution operation.
The loss function of $ l_e $ is similar to Eqn.  \ref{Eq.L_bce} where the saliency prediction and ground-truth are replaced with edge prediction and ground-truth.

The whole training loss $ L $ of our network is the sum of $ L_S $ in Eqn.\ref{Eq.S_spvs} and $ L_E $ in Eqn.\ref{Eq.E_spvs}:
\begin{equation}\label{Eq.L_sum} L  = (L_S+L_E)/2  \end{equation}

\begin{figure*}
	\begin{center}
		\includegraphics[width=1\textwidth]{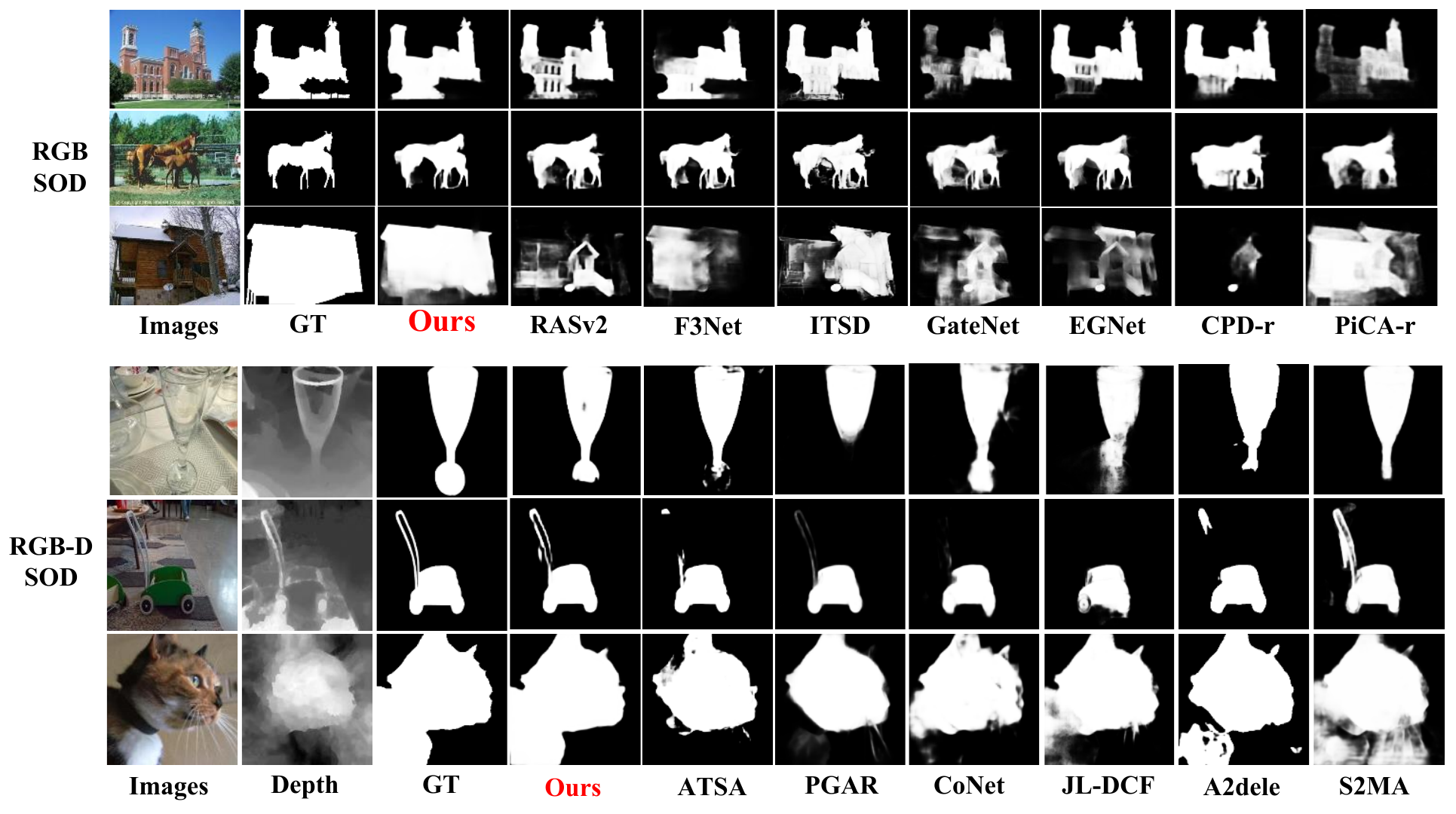}
	\end{center}
	\caption{Visual comparisons of the proposed method and state-of-the-art methods. Our method obtains more accurate results than others in both tasks of RGB and RGBD image SOD.}
	\label{fig:VC}
\end{figure*}

\begin{figure*}
	\begin{center}
		\includegraphics[width=0.8\textwidth]{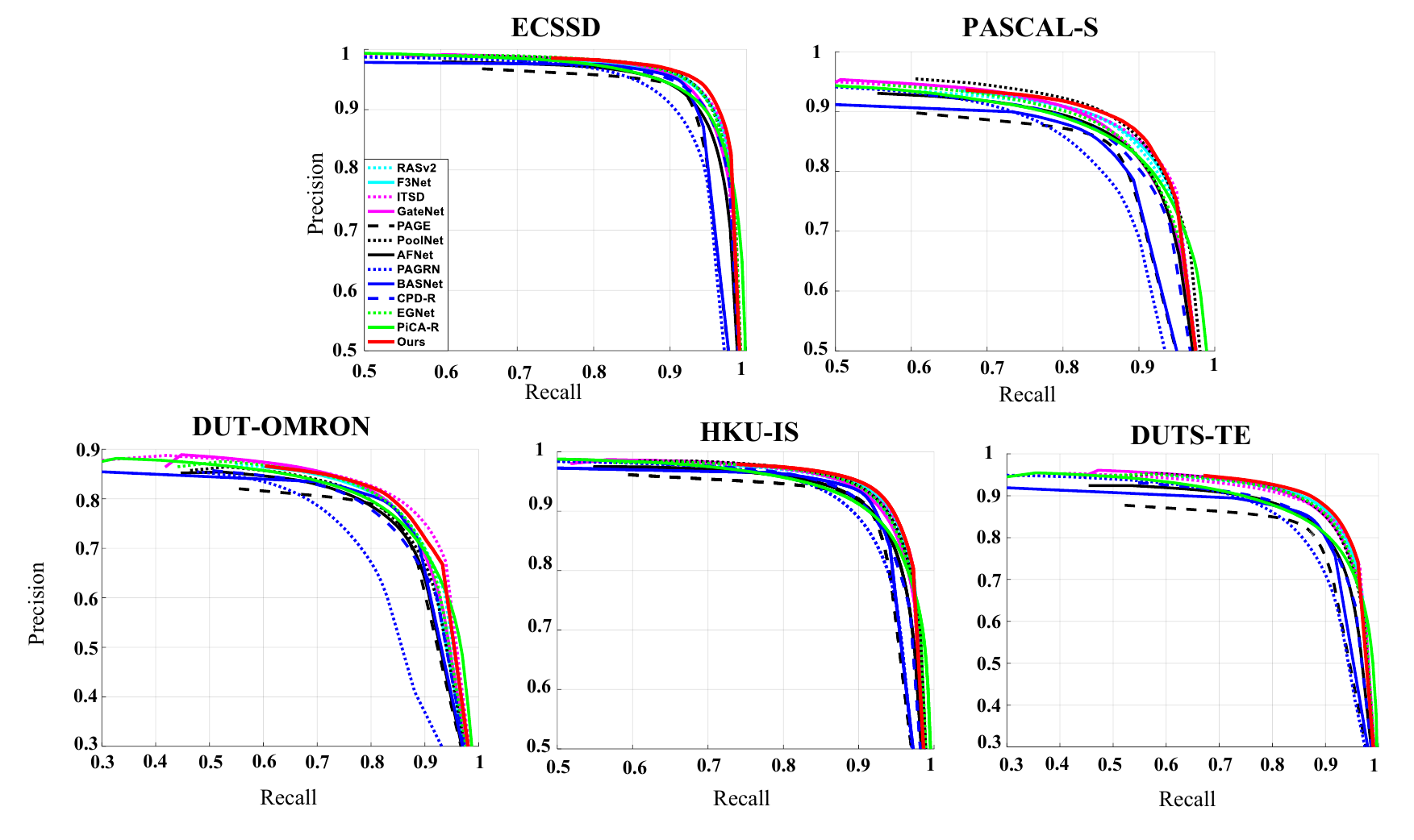}
	\end{center}
	\caption{Comparison of precision--recall curves. Our method obtains the best performance on four datasets and the second best performance on DUT-OMRON.}
	\label{fig_rgb_pr}
\end{figure*}

\begin{table*}[t]
	\centering
	\caption{Quantitative comparisons of RGB image SOD methods on five datasets in terms of six metrics. The best results are adopted from the original work, if provided: for example, CPD-r \cite{wu2019cascaded} and PiCA-r \cite{liu2018picanet} are the results of ResNet versions, and PAGE\_crf \cite{wang2019PAGE} is the result after post-processing. Values in {\color{red}{\textbf{red}}} and  {\color{green}{\textbf{green}}} indicate the best and second best results.}
	\resizebox{\textwidth}{55mm}{
		\begin{tabular}{|c|l|c|cccccccccccc|}
			\toprule
			\multicolumn{2}{|c|}{\textbf{Publish Date}} & \multicolumn{1}{c|}{\multirow{2}[4]{*}{\textbf{Ours}}} & \multicolumn{1}{c|}{AAAI'20} & \multicolumn{1}{c|}{TIP'20} & \multicolumn{1}{c|}{CVPR'20} & \multicolumn{1}{c|}{ECCV'20} & \multicolumn{1}{c|}{ICCV'20} & \multicolumn{5}{c|}{CVPR'19}          & \multicolumn{2}{c|}{CVPR'18} \\
			\cmidrule{1-2}
			\cmidrule{4-15}    
			\textbf{Dataset} & \textbf{Metric} & \multicolumn{1}{c|}{} & \multicolumn{1}{c|}{\textbf{F3Net}} & \multicolumn{1}{c|}{\textbf{RASv2}} & \multicolumn{1}{c|}{\textbf{ITSD}} & \multicolumn{1}{c|}{\textbf{GateNet}} & \multicolumn{1}{c|}{\textbf{EGNet}} & \textbf{PoolNet} & \textbf{BASNet} & \textbf{CPD-r} & \textbf{PAGE\_crf} & \multicolumn{1}{c|}{\textbf{AFNet}} & \textbf{PiCA-r} & \textbf{PAGRN} \\
			\midrule
			\multirow{6}[1]{*}{\textbf{ECSSD}} & \multicolumn{1}{l|}{maxF $ \uparrow $} & \textcolor[rgb]{ 1,  0,  0}{\textbf{0.951}} & 0.945 & 0.948 & 0.947 & 0.945 & 0.947 & \textcolor[rgb]{ 0,  .69,  .314}{\textbf{0.949}} & 0.942 & 0.939 & 0.932 & 0.935 & 0.935 & 0.927 \\
			& \multicolumn{1}{l|}{mF $ \uparrow $} & \textcolor[rgb]{ 0,  .69,  .314}{\textbf{0.924}} & \textcolor[rgb]{ 1,  0,  0}{\textbf{0.925}} & 0.923 & 0.895 & 0.916 & 0.92  & 0.919 & 0.88  & 0.917 & 0.915 & 0.908 & 0.886 & 0.894 \\
			& \multicolumn{1}{l|}{MAE $ \downarrow $} & \textcolor[rgb]{ 1,  0,  0}{\textbf{0.032}} & \textcolor[rgb]{ 0,  .69,  .314}{\textbf{0.033}} & 0.034 & 0.034 & 0.04  & 0.037 & 0.035 & 0.037 & 0.037 & 0.04  & 0.042 & 0.046 & 0.061 \\
			& \multicolumn{1}{l|}{wF $ \uparrow $} & \textcolor[rgb]{ 1,  0,  0}{\textbf{0.918}} & 0.912 & \textcolor[rgb]{ 0,  .69,  .314}{\textbf{0.913}} & 0.91  & 0.894 & 0.903 & 0.904 & 0.904 & 0.898 & 0.895 & 0.886 & 0.867 & 0.833 \\
			& \multicolumn{1}{l|}{Sm $ \uparrow $} & \textcolor[rgb]{ 1,  0,  0}{\textbf{0.93}} & 0.924 & 0.925 & 0.925 & 0.92  & 0.925 & \textcolor[rgb]{ 0,  .69,  .314}{\textbf{0.926}} & 0.916 & 0.918 & 0.911 & 0.913 & 0.917 & 0.889 \\
			& \multicolumn{1}{l|}{Em $ \uparrow $} & 0.922 & 0.927 & \textcolor[rgb]{ 1,  0,  0}{\textbf{0.928}} & \textcolor[rgb]{ 0,  .69,  .314}{\textbf{0.927}} & 0.924 & \textcolor[rgb]{ 0,  .69,  .314}{\textbf{0.927}} & 0.925 & 0.921 & 0.925 & 0.921 & 0.918 & 0.913 & 0.914 \\
			\midrule
			
			\multicolumn{1}{|c|}{\multirow{6}[0]{*}{\textbf{PASCAL\newline{}\_S}}} & \multicolumn{1}{l|}{maxF} & \textcolor[rgb]{ 0,  .69,  .314}{\textbf{0.889}} & 0.882 & 0.881 & 0.882 & 0.881 & 0.876 & \textcolor[rgb]{ 1,  0,  0}{\textbf{0.892}} & 0.86  & 0.87  & 0.856 & 0.871 & 0.868 & 0.855 \\
			& \multicolumn{1}{l|}{mF} & \textcolor[rgb]{ 1,  0,  0}{\textbf{0.849}} & \textcolor[rgb]{ 0,  .69,  .314}{\textbf{0.844}} & 0.836 & 0.797 & 0.83  & 0.829 & 0.838 & 0.777 & 0.829 & 0.823 & 0.828 & 0.802 & 0.805 \\
			& \multicolumn{1}{l|}{MAE} & \textcolor[rgb]{ 1,  0,  0}{\textbf{0.064}} & \textcolor[rgb]{ 1,  0,  0}{\textbf{0.064}} & 0.067 & \textcolor[rgb]{ 0,  .69,  .314}{\textbf{0.066}} & 0.071 & 0.076 & 0.067 & 0.079 & 0.074 & 0.077 & 0.071 & 0.078 & 0.095 \\
			& \multicolumn{1}{l|}{wF} & \textcolor[rgb]{ 1,  0,  0}{\textbf{0.83}} & \textcolor[rgb]{ 0,  .69,  .314}{\textbf{0.823}} & 0.82  & \textcolor[rgb]{ 0,  .69,  .314}{\textbf{0.823}} & 0.804 & 0.804 & 0.819 & 0.797 & 0.8   & 0.796 & 0.804 & 0.779 & 0.734 \\
			& \multicolumn{1}{l|}{Sm} & \textcolor[rgb]{ 0,  .69,  .314}{\textbf{0.863}} & 0.857 & 0.855 & 0.859 & 0.854 & 0.85  & \textcolor[rgb]{ 1,  0,  0}{\textbf{0.864}} & 0.834 & 0.844 & 0.834 & 0.85  & 0.852 & 0.814 \\
			& \multicolumn{1}{l|}{Em} & 0.856 & \textcolor[rgb]{ 1,  0,  0}{\textbf{0.863}} & 0.855 & \textcolor[rgb]{ 0,  .69,  .314}{\textbf{0.859}} & 0.856 & 0.854 & \textcolor[rgb]{ 0,  .69,  .314}{\textbf{0.859}} & 0.85  & 0.853 & 0.848 & 0.853 & 0.837 & 0.848 \\
			\midrule
			
			\multicolumn{1}{|c|}{\multirow{6}[0]{*}{\textbf{DUTS\newline{}-TEST}}} & \multicolumn{1}{l|}{maxF} & \textcolor[rgb]{ 1,  0,  0}{\textbf{0.896}} & \textcolor[rgb]{ 0,  .69,  .314}{\textbf{0.891}} & 0.886 & 0.883 & 0.888 & 0.889 & 0.889 & 0.859 & 0.865 & 0.839 & 0.863 & 0.86  & 0.854 \\
			& \multicolumn{1}{l|}{mF} & \textcolor[rgb]{ 1,  0,  0}{\textbf{0.857}} & 0.84  & 0.831 & 0.804 & 0.807 & 0.815 & 0.819 & 0.791 & 0.805 & 0.796 & 0.792 & 0.759 & 0.784 \\
			& \multicolumn{1}{l|}{MAE} & \textcolor[rgb]{ 0,  .69,  .314}{\textbf{0.036}} & \textcolor[rgb]{ 1,  0,  0}{\textbf{0.035}} & 0.037 & 0.041 & 0.04  & 0.039 & 0.037 & 0.048 & 0.043 & 0.05  & 0.046 & 0.051 & 0.056 \\
			& \multicolumn{1}{l|}{wF} & \textcolor[rgb]{ 1,  0,  0}{\textbf{0.844}} & \textcolor[rgb]{ 0,  .69,  .314}{\textbf{0.835}} & 0.827 & 0.823 & 0.809 & 0.815 & 0.817 & 0.803 & 0.795 & 0.779 & 0.785 & 0.754 & 0.724 \\
			& \multicolumn{1}{l|}{Sm} & \textcolor[rgb]{ 1,  0,  0}{\textbf{0.892}} & \textcolor[rgb]{ 0,  .69,  .314}{\textbf{0.888}} & 0.884 & 0.859 & 0.885 & 0.887 & 0.886 & 0.866 & 0.869 & 0.851 & 0.867 & 0.869 & 0.838 \\
			& \multicolumn{1}{l|}{Em} & \textcolor[rgb]{ 1,  0,  0}{\textbf{0.908}} & \textcolor[rgb]{ 0,  .69,  .314}{\textbf{0.902}} & 0.901 & 0.895 & 0.889 & 0.891 & 0.896 & 0.884 & 0.886 & 0.873 & 0.879 & 0.862 & 0.88 \\
			\midrule
			
			\multirow{6}[0]{*}{\textbf{HKU-IS}} & \multicolumn{1}{l|}{maxF} & \textcolor[rgb]{ 1,  0,  0}{\textbf{0.939}} & \textcolor[rgb]{ 0,  .69,  .314}{\textbf{0.937}} & 0.933 & 0.934 & 0.933 & 0.935 & 0.936 & 0.928 & 0.925 & 0.919 & 0.923 & 0.918 & 0.918 \\
			& \multicolumn{1}{l|}{mF} & \textcolor[rgb]{ 1,  0,  0}{\textbf{0.918}} & \textcolor[rgb]{ 0,  .69,  .314}{\textbf{0.91}} & 0.906 & 0.899 & 0.899 & 0.901 & 0.903 & 0.895 & 0.891 & 0.895 & 0.888 & 0.87  & 0.886 \\
			& \multicolumn{1}{l|}{MAE} & \textcolor[rgb]{ 1,  0,  0}{\textbf{0.027}} & \textcolor[rgb]{ 0,  .69,  .314}{\textbf{0.028}} & 0.03  & 0.031 & 0.033 & 0.031 & 0.03  & 0.032 & 0.034 & 0.034 & 0.036 & 0.043 & 0.048 \\
			& \multicolumn{1}{l|}{wF} & \textcolor[rgb]{ 1,  0,  0}{\textbf{0.907}} & \textcolor[rgb]{ 0,  .69,  .314}{\textbf{0.9}} & 0.894 & 0.894 & 0.88  & 0.887 & 0.888 & 0.889 & 0.875 & 0.877 & 0.869 & 0.84  & 0.82 \\
			& \multicolumn{1}{l|}{Sm} & \textcolor[rgb]{ 1,  0,  0}{\textbf{0.922}} & 0.917 & 0.915 & 0.917 & 0.915 & 0.918 & \textcolor[rgb]{ 0,  .69,  .314}{\textbf{0.919}} & 0.909 & 0.905 & 0.902 & 0.905 & 0.904 & 0.887 \\
			& \multicolumn{1}{l|}{Em} & \textcolor[rgb]{ 1,  0,  0}{\textbf{0.955}} & \textcolor[rgb]{ 0,  .69,  .314}{\textbf{0.953}} & 0.952 & 0.952 & 0.949 & 0.95  & \textcolor[rgb]{ 0,  .69,  .314}{\textbf{0.953}} & 0.946 & 0.944 & 0.941 & 0.942 & 0.936 & 0.939 \\
			\midrule
			
			\multicolumn{1}{|c|}{\multirow{6}[1]{*}{\textbf{DUT-OMRON}}} & \multicolumn{1}{l|}{maxF} & \textcolor[rgb]{ 0,  .69,  .314}{\textbf{0.819}} & 0.813 & 0.815 & \textcolor[rgb]{ 1,  0,  0}{\textbf{0.821}} & 0.818 & 0.815 & 0.805 & 0.805 & 0.797 & 0.792 & 0.797 & 0.803 & 0.771 \\
			& \multicolumn{1}{l|}{mF} & \textcolor[rgb]{ 1,  0,  0}{\textbf{0.776}} & \textcolor[rgb]{ 0,  .69,  .314}{\textbf{0.766}} & 0.763 & 0.756 & 0.746 & 0.755 & 0.752 & 0.756 & 0.747 & 0.747 & 0.738 & 0.717 & 0.711 \\
			& \multicolumn{1}{l|}{MAE} & 0.058 & \textcolor[rgb]{ 1,  0,  0}{\textbf{0.053}} & 0.055 & 0.061 & 0.055 & \textcolor[rgb]{ 1,  0,  0}{\textbf{0.053}} & \textcolor[rgb]{ 0,  .69,  .314}{\textbf{0.054}} & 0.056 & 0.056 & 0.061 & 0.057 & 0.065 & 0.071 \\
			& \multicolumn{1}{l|}{wF} & \textcolor[rgb]{ 1,  0,  0}{\textbf{0.752}} & \textcolor[rgb]{ 0,  .69,  .314}{\textbf{0.747}} & 0.743 & 0.75  & 0.729 & 0.738 & 0.725 & 0.751 & 0.719 & 0.729 & 0.717 & 0.695 & 0.622 \\
			& \multicolumn{1}{l|}{Sm} & \textcolor[rgb]{ 1,  0,  0}{\textbf{0.841}} & 0.838 & 0.836 & \textcolor[rgb]{ 0,  .69,  .314}{\textbf{0.84}} & 0.838 & \textcolor[rgb]{ 1,  0,  0}{\textbf{0.841}} & 0.831 & 0.836 & 0.825 & 0.822 & 0.826 & 0.832 & 0.775 \\
			& Em    & \textcolor[rgb]{ 1,  0,  0}{\textbf{0.873}} & 0.87  & \textcolor[rgb]{ 0,  .69,  .314}{\textbf{0.872}} & 0.863 & 0.862 & 0.867 & 0.868 & 0.869 & 0.866 & 0.851 & 0.853 & 0.841 & 0.842 \\
			\bottomrule
	\end{tabular}}
	\label{tab:rgb}%
\end{table*}%

\begin{table*}[t]
	\centering
	\caption{Quantitative comparisons of RGB-D image SOD methods on seven datasets in terms of six metrics. The best results are compared if authors provided results or they could be generated from released code with recommended configurations. Values in {\color{red}{\textbf{red}}} and  {\color{green}{\textbf{green}}} indicate the best and second best results.}
	\resizebox{\textwidth}{70mm}{
		\begin{tabular}{|c|l|ccccccccccccc|}
			\toprule
			\multicolumn{2}{|c|}{\textbf{Publish Date}} & \multicolumn{1}{c|}{\multirow{2}[4]{*}{\textbf{Ours}}} & \multicolumn{3}{c|}{ECCV'20} & \multicolumn{5}{c|}{CVPR'20}          & \multicolumn{1}{c|}{TNNLS'20} & \multicolumn{1}{c|}{TIP'20} & \multicolumn{1}{c|}{ICCV'19} & CVPR'19 \\
			\cmidrule{1-2} \cmidrule{4-15}    \textbf{Dataset} & \multicolumn{1}{c|}{\textbf{Metric}} & \multicolumn{1}{c|}{} & \textbf{ATSA} & \textbf{PGAR} & \multicolumn{1}{c|}{\textbf{CoNet}} & \textbf{JL-DCF} & \textbf{A2dele} & \textbf{S2MA} & \textbf{UCNet} & \multicolumn{1}{c|}{\textbf{SSF}} & \multicolumn{1}{c|}{\textbf{D3Net}} & \multicolumn{1}{c|}{\textbf{ICNet}} & \multicolumn{1}{c|}{\textbf{DMRA}} & \textbf{CPFP} \\
			\midrule
			\multirow{6}[2]{*}{\textbf{DUT-RGBD}} & maxF $ \uparrow $ & \textcolor[rgb]{ 1,  0,  0}{\textbf{0.946}} & \textcolor[rgb]{ 0,  .69,  .314}{\textbf{0.943}} & 0.938 & 0.936 & 0.923 & 0.907 & 0.91  & 0.883 & 0.934 & 0.771 & 0.874 & 0.909 & 0.571 \\
			& mF $ \uparrow $   & \textcolor[rgb]{ 1,  0,  0}{\textbf{0.933}} & \textcolor[rgb]{ 0,  .69,  .314}{\textbf{0.919}} & 0.913 & 0.909 & 0.883 & 0.891 & 0.886 & 0.857 & 0.915 & 0.739 & 0.83  & 0.884 & 0.514 \\
			& MAE  $ \downarrow $ & \textcolor[rgb]{ 1,  0,  0}{\textbf{0.027}} & \textcolor[rgb]{ 0,  .69,  .314}{\textbf{0.031}} & 0.035 & 0.033 & 0.043 & 0.041 & 0.043 & 0.056 & 0.033 & 0.099 & 0.072 & 0.048 & 0.148 \\
			& wF  $ \uparrow $  & \textcolor[rgb]{ 1,  0,  0}{\textbf{0.92}} & \textcolor[rgb]{ 0,  .69,  .314}{\textbf{0.908}} & 0.893 & 0.896 & 0.868 & 0.87  & 0.868 & 0.828 & 0.9   & 0.668 & 0.79  & 0.858 & 0.412 \\
			& Sm  $ \uparrow $  & \textcolor[rgb]{ 1,  0,  0}{\textbf{0.927}} & \textcolor[rgb]{ 0,  .69,  .314}{\textbf{0.921}} & 0.919 & 0.919 & 0.906 & 0.88  & 0.903 & 0.863 & 0.915 & 0.775 & 0.852 & 0.888 & 0.605 \\
			& Em  $ \uparrow $  & \textcolor[rgb]{ 1,  0,  0}{\textbf{0.953}} & 0.946 & 0.944 & \textcolor[rgb]{ 0,  .69,  .314}{\textbf{0.948}} & 0.931 & 0.921 & 0.921 & 0.903 & 0.946 & 0.85  & 0.897 & 0.927 & 0.685 \\
			\midrule
			\multirow{6}[2]{*}{\textbf{LFSD}} & maxF  & \textcolor[rgb]{ 1,  0,  0}{\textbf{0.894}} & 0.884 & 0.874 & 0.877 & 0.887 & 0.857 & 0.862 & 0.878 & \textcolor[rgb]{ 0,  .69,  .314}{\textbf{0.893}} & 0.849 & 0.891 & 0.872 & 0.849 \\
			& mF    & \textcolor[rgb]{ 1,  0,  0}{\textbf{0.874}} & 0.862 & 0.852 & 0.848 & 0.854 & 0.835 & 0.803 & 0.859 & \textcolor[rgb]{ 0,  .69,  .314}{\textbf{0.867}} & 0.792 & 0.861 & 0.849 & 0.813 \\
			& MAE   & \textcolor[rgb]{ 1,  0,  0}{\textbf{0.06}} & \textcolor[rgb]{ 0,  .69,  .314}{\textbf{0.064}} & 0.074 & 0.071 & 0.07  & 0.074 & 0.095 & 0.066 & 0.066 & 0.099 & 0.071 & 0.075 & 0.088 \\
			& wF    & \textcolor[rgb]{ 1,  0,  0}{\textbf{0.847}} & \textcolor[rgb]{ 0,  .69,  .314}{\textbf{0.841}} & 0.803 & 0.819 & 0.825 & 0.81  & 0.774 & 0.834 & 0.831 & 0.759 & 0.825 & 0.814 & 0.778 \\
			& Sm    & \textcolor[rgb]{ 0,  .69,  .314}{\textbf{0.866}} & 0.865 & 0.853 & 0.862 & 0.862 & 0.837 & 0.837 & 0.865 & 0.859 & 0.832 & \textcolor[rgb]{ 1,  0,  0}{\textbf{0.868}} & 0.847 & 0.828 \\
			& Em    & \textcolor[rgb]{ 1,  0,  0}{\textbf{0.908}} & \textcolor[rgb]{ 0,  .69,  .314}{\textbf{0.905}} & 0.889 & 0.896 & 0.882 & 0.87  & 0.863 & 0.897 & 0.895 & 0.833 & 0.891 & 0.899 & 0.867 \\
			\midrule
			\multirow{6}[1]{*}{\textbf{NJUD}} & maxF  & \textcolor[rgb]{ 1,  0,  0}{\textbf{0.923}} & \textcolor[rgb]{ 0,  .69,  .314}{\textbf{0.917}} & \textcolor[rgb]{ 0,  .69,  .314}{\textbf{0.917}} & 0.902 & 0.911 & 0.889 & 0.899 & 0.907 & 0.911 & 0.831 & 0.901 & 0.897 & 0.89 \\
			& mF    & \textcolor[rgb]{ 1,  0,  0}{\textbf{0.905}} & \textcolor[rgb]{ 0,  .69,  .314}{\textbf{0.893}} & 0.892 & 0.873 & 0.883 & 0.872 & 0.836 & 0.888 & 0.886 & 0.784 & 0.867 & 0.872 & 0.836 \\
			& MAE   & \textcolor[rgb]{ 1,  0,  0}{\textbf{0.037}} & \textcolor[rgb]{ 0,  .69,  .314}{\textbf{0.04}} & 0.042 & 0.046 & 0.042 & 0.049 & 0.054 & 0.043 & 0.043 & 0.076 & 0.052 & 0.051 & 0.053 \\
			& wF    & \textcolor[rgb]{ 1,  0,  0}{\textbf{0.892}} & \textcolor[rgb]{ 0,  .69,  .314}{\textbf{0.883}} & 0.876 & 0.856 & 0.874 & 0.851 & 0.846 & 0.872 & 0.871 & 0.747 & 0.849 & 0.853 & 0.834 \\
			& Sm    & \textcolor[rgb]{ 0,  .69,  .314}{\textbf{0.907}} & 0.901 & \textcolor[rgb]{ 1,  0,  0}{\textbf{0.909}} & 0.895 & 0.903 & 0.867 & 0.894 & 0.896 & 0.899 & 0.827 & 0.894 & 0.886 & 0.878 \\
			& Em    & \textcolor[rgb]{ 1,  0,  0}{\textbf{0.926}} & \textcolor[rgb]{ 0,  .69,  .314}{\textbf{0.921}} & 0.916 & 0.912 & 0.912 & 0.896 & 0.896 & 0.903 & 0.913 & 0.873 & 0.904 & 0.908 & 0.895 \\
			
			\midrule
			\multirow{6}[1]{*}{\textbf{NLPR}} & maxF  & 0.919 & 0.908 & \textcolor[rgb]{ 0,  .69,  .314}{\textbf{0.925}} & 0.898 & \textcolor[rgb]{ 1,  0,  0}{\textbf{0.926}} & 0.896 & 0.91  & 0.915 & 0.912 & 0.905 & 0.919 & 0.888 & 0.888 \\
			& mF    & \textcolor[rgb]{ 1,  0,  0}{\textbf{0.898}} & 0.869 & 0.883 & 0.848 & 0.875 & 0.878 & 0.847 & \textcolor[rgb]{ 0,  .69,  .314}{\textbf{0.886}} & 0.875 & 0.833 & 0.867 & 0.854 & 0.821 \\
			& MAE   & \textcolor[rgb]{ 0,  .69,  .314}{\textbf{0.023}} & 0.028 & 0.025 & 0.031 & \textcolor[rgb]{ 1,  0,  0}{\textbf{0.022}} & 0.028 & 0.03  & 0.025 & 0.026 & 0.034 & 0.028 & 0.031 & 0.036 \\
			& wF    & \textcolor[rgb]{ 1,  0,  0}{\textbf{0.894}} & 0.863 & 0.886 & 0.85  & \textcolor[rgb]{ 0,  .69,  .314}{\textbf{0.887}} & 0.867 & 0.857 & 0.883 & 0.874 & 0.833 & 0.87  & 0.845 & 0.82 \\
			& Sm    & 0.921 & 0.905 & \textcolor[rgb]{ 1,  0,  0}{\textbf{0.93}} & 0.908 & \textcolor[rgb]{ 0,  .69,  .314}{\textbf{0.925}} & 0.896 & 0.915 & 0.92  & 0.914 & 0.906 & 0.922 & 0.899 & 0.888 \\
			& Em    & \textcolor[rgb]{ 1,  0,  0}{\textbf{0.956}} & 0.942 & \textcolor[rgb]{ 0,  .69,  .314}{\textbf{0.954}} & 0.934 & 0.952 & 0.945 & 0.937 & 0.952 & 0.949 & 0.931 & 0.943 & 0.941 & 0.924 \\
			\midrule
			\multirow{6}[1]{*}{\textbf{RGBD135}} & maxF  & \textcolor[rgb]{ 0,  .69,  .314}{\textbf{0.938}} & 0.937 & 0.929 & 0.916 & 0.936 & 0.89  & \textcolor[rgb]{ 1,  0,  0}{\textbf{0.946}} & 0.937 & 0.912 & 0.916 & 0.928 & 0.906 & 0.883 \\
			& mF    & \textcolor[rgb]{ 0,  .69,  .314}{\textbf{0.902}} & \textcolor[rgb]{ 0,  .69,  .314}{\textbf{0.902}} & 0.869 & 0.862 & 0.886 & 0.864 & 0.893 & \textcolor[rgb]{ 1,  0,  0}{\textbf{0.906}} & 0.876 & 0.862 & 0.877 & 0.857 & 0.818 \\
			& MAE   & \textcolor[rgb]{ 0,  .69,  .314}{\textbf{0.02}} & \textcolor[rgb]{ 1,  0,  0}{\textbf{0.018}} & 0.025 & 0.027 & \textcolor[rgb]{ 0,  .69,  .314}{\textbf{0.02}} & 0.027 & \textcolor[rgb]{ 0,  .69,  .314}{\textbf{0.02}} & \textcolor[rgb]{ 1,  0,  0}{\textbf{0.018}} & 0.025 & 0.03  & 0.026 & 0.029 & 0.037 \\
			& wF    & 0.893 & \textcolor[rgb]{ 0,  .69,  .314}{\textbf{0.908}} & 0.862 & 0.856 & 0.897 & 0.845 & 0.897 & \textcolor[rgb]{ 1,  0,  0}{\textbf{0.913}} & 0.86  & 0.837 & 0.875 & 0.849 & 0.794 \\
			& Sm    & 0.92  & 0.928 & 0.916 & 0.911 & 0.931 & 0.878 & \textcolor[rgb]{ 1,  0,  0}{\textbf{0.943}} & \textcolor[rgb]{ 0,  .69,  .314}{\textbf{0.935}} & 0.905 & 0.906 & 0.923 & 0.901 & 0.874 \\
			& Em    & 0.953 & 0.969 & 0.94  & 0.945 & 0.966 & 0.919 & \textcolor[rgb]{ 0,  .69,  .314}{\textbf{0.971}} & \textcolor[rgb]{ 1,  0,  0}{\textbf{0.973}} & 0.948 & 0.953 & 0.958 & 0.945 & 0.927 \\
			
			\midrule
			\multirow{6}[1]{*}{\textbf{SSD}} & maxF  & \textcolor[rgb]{ 0,  .69,  .314}{\textbf{0.874}} & 0.873 & 0.85  & 0.851 & 0.799 & 0.815 & \textcolor[rgb]{ 1,  0,  0}{\textbf{0.878}} & 0.723 & 0.838 & 0.872 & 0.855 & 0.858 & 0.801 \\
			& mF    & \textcolor[rgb]{ 1,  0,  0}{\textbf{0.844}} & \textcolor[rgb]{ 0,  .69,  .314}{\textbf{0.827}} & 0.799 & 0.806 & 0.768 & 0.79  & 0.818 & 0.693 & 0.799 & 0.793 & 0.798 & 0.821 & 0.725 \\
			& MAE   & \textcolor[rgb]{ 1,  0,  0}{\textbf{0.048}} & \textcolor[rgb]{ 0,  .69,  .314}{\textbf{0.05}} & 0.066 & 0.059 & 0.071 & 0.068 & 0.053 & 0.1   & 0.061 & 0.058 & 0.064 & 0.058 & 0.082 \\
			& wF    & \textcolor[rgb]{ 1,  0,  0}{\textbf{0.822}} & \textcolor[rgb]{ 0,  .69,  .314}{\textbf{0.814}} & 0.769 & 0.792 & 0.731 & 0.753 & 0.794 & 0.636 & 0.772 & 0.792 & 0.781 & 0.797 & 0.718 \\
			& Sm    & \textcolor[rgb]{ 0,  .69,  .314}{\textbf{0.863}} & 0.86  & 0.844 & 0.853 & 0.814 & 0.802 & \textcolor[rgb]{ 1,  0,  0}{\textbf{0.868}} & 0.741 & 0.837 & 0.867 & 0.848 & 0.857 & 0.807 \\
			& Em    & \textcolor[rgb]{ 1,  0,  0}{\textbf{0.91}} & \textcolor[rgb]{ 0,  .69,  .314}{\textbf{0.901}} & 0.872 & 0.896 & 0.886 & 0.858 & 0.891 & 0.844 & 0.887 & 0.887 & 0.878 & 0.892 & 0.832 \\
			\midrule
			\multirow{6}[2]{*}{\textbf{STEREO}} & maxF  & \textcolor[rgb]{ 1,  0,  0}{\textbf{0.915}} & 0.909 & \textcolor[rgb]{ 0,  .69,  .314}{\textbf{0.914}} & 0.912 & 0.89  & 0.896 & 0.835 & 0.831 & 0.902 & 0.862 &  -    & 0.895 & 0.737 \\
			& mF    & \textcolor[rgb]{ 1,  0,  0}{\textbf{0.899}} & 0.884 & 0.885 & 0.885 & 0.871 & 0.883 & \textcolor[rgb]{ 0,  .69,  .314}{\textbf{0.896}} & 0.811 & 0.88  & 0.82  & -   & 0.868 & 0.696 \\
			& MAE   & \textcolor[rgb]{ 1,  0,  0}{\textbf{0.038}} & \textcolor[rgb]{ 0,  .69,  .314}{\textbf{0.039}} & 0.044 & 0.04  & 0.051 & 0.042 & 0.078 & 0.073 & 0.044 & 0.062 &   -  & 0.047 & 0.097 \\
			& wF    & \textcolor[rgb]{ 1,  0,  0}{\textbf{0.885}} & \textcolor[rgb]{ 0,  .69,  .314}{\textbf{0.874}} & 0.86  & 0.871 & 0.837 & 0.867 & 0.758 & 0.774 & 0.862 & 0.785 &  -  & 0.85  & 0.629 \\
			& Sm    & \textcolor[rgb]{ 0,  .69,  .314}{\textbf{0.902}} & 0.897 & 0.901 & \textcolor[rgb]{ 1,  0,  0}{\textbf{0.908}} & 0.885 & 0.883 & 0.838 & 0.829 & 0.893 & 0.857 &   -  & 0.886 & 0.741 \\
			& Em    & \textcolor[rgb]{ 1,  0,  0}{\textbf{0.932}} & 0.921 & 0.922 & \textcolor[rgb]{ 0,  .69,  .314}{\textbf{0.924}} & 0.917 & 0.915 & 0.868 & 0.881 & 0.923 & 0.898 &  -   & 0.92  & 0.809 \\
			\bottomrule
	\end{tabular}}
	\label{tab:rgbd}%
\end{table*}%

\subsection {RGB-D Salient Object Detection}
With a slight adaption, our structure can be directly used for RGB-D SOD.
Compared to RGB image SOD, there is one more input (depth) in RGB-D image SOD.
Therefore, we add depth feature inputs via a backbone network, as shown in Fig.\ref{fig_unified} .

\paragraph{CRACE with Depth Flow}
Since the depth features are added into the network, the CRACE module formulated by Eqn.\ref{Eq.decoder} is slightly adjusted for the task of RGB-D image SOD:
\begin{equation}\label{Eq.rgbd-decoder} F_{i-1} =  CRACE(f_{i-1}, d_{i-1}, F_i )  \end{equation}
where $ d_i $ represents the depth feature generated from the backbone network.

\paragraph{Supervision and Loss}
Following the design of multi-level supervision in the previous section, we add 4-level saliency supervision for the depth context. For simplicity, we do not apply an edge supervision to the depth.
\begin{equation}\label{Eq.L_sum_rgbd} L_{rgbd}  = (L_S+L_D+L_E) /3\end{equation}
where $ L_D $ is the same loss function as in Eqn.\ref{Eq.S_spvs} but the prediction map is generated by depth.

\section{Experiments}
To thoroughly evaluate our model, we conducted a series of experiments for both RGB and RGB-D image SOD tasks on various popular datasets in terms of seven widely used metrics.
Ablation experiments and analyses are also conducted to exhaustively validate the effectiveness of the proposed module and framework.
Some failure cases are also listed to provide direction for further improvement.

\subsection{Implementation Details}
Our network is trained on the DUTS-TR dataset, like many other methods \cite{wang2017learning}; this dataset contains 10553 images.
We use the ResNet50 \cite{he2016deep} pre-trained on ImageNet \cite{deng2009imagenet} as a backbone network.
The input image is first resized to 352$\times$352 pixels.
The training strategy is the same as for RASv2 \cite{chen2020reverse} and F3Net \cite{wei2019f3net}.
In this stage, we adopted the stochastic gradient descent (SGD) with a momentum of 0.9 and a weight decay set to 0.0005.
The initial learning rate for the backbone is 0.005 and that for the head is 0.05. Warm-up processing is applied to adjust the learning rate during training.   
The horizontal flip, random crop, and multi-scale pre-processing are adopted for data augmentation.
Our model is written in Pytorch \cite{paszke2019pytorch} and runs on an NVIDIA 1080ti GPU.
The whole training process takes about 7.5 hours with a batch size of 32.
The inference speed for an image is about 37 fps.

For training for the RGB-D task, we follow \cite{Piao2020A2deleAA} and adopt their training set, which contains 800 samples from the DUT-RGBD dataset, 1485 samples from the NJUD dataset, and 700 samples from the NLPR dataset. 
The learning rate of the depth branch is the same as that of the RGB decoder in our network.
All other settings are the same as those used in RGB image SOD.

\subsection{Metrics}
Both tasks can be evaluated in terms of the same metrics. We use the precision--recall (PR) curve, maximum F-measure ($ maxF $), mean F-measure ($ mF $), weighted F-score ($wF$), mean absolute error ($ MAE $), structure measure ($S_m$), and  enhanced-alignment measure ($ E_m $).
Precision and recall are computed by comparing the threshed saliency maps with the ground-truth maps. 
Usually, precision and recall values are not suitable for the evaluation because precision is more important than recall \cite{liu2010learning}.
Consequently, $F_\beta$ is proposed to solve this problem. The F-measure is defined as
\begin{equation}\label{Eq.F_beta} F_\beta = \frac{(1+\beta ^2)\cdot Precision\cdot Recall}{\beta ^2\cdot Precision + Recall} \end{equation}
where $\beta^2$ is set to 0.3, as suggested by \cite{achanta2009frequency,ChengPAMI}.
Metrics $ maxF $ and $ mF $ are the maximum and mean values of the F-measure.
$ MAE $ is the mean absolute error between saliency maps and ground-truth maps \cite{Hornung2012Saliency}.
The weighted F-measure ($ wF $) \cite{margolin2014evaluate} is utilized since it is believed to be more reliable by eliminating the interpolation, dependency, and equal importance flaws of the previously described measures.
Recently, new metrics have been proposed: the structure measure ($ S_m $) \cite{fan2017structure} and enhanced-alignment measure ($ E_m $)\cite{fan2018enhanced}. These are believed to be better for measuring the structural similarity and image-pixel level matching information between saliency maps and ground-truth maps.

\subsection{RGB image SOD}
\paragraph{Datasets}
Five popular SOD datasets are used: ECSSD \cite{shi2016hierarchical} with 1000 structurally complex images, PASCAL-S \cite{li2014secrets} with 850 images with cluttered backgrounds, DUT-OMRON \cite{yang2013saliency} with 5168 multi-object images, HKU-IS \cite{li2015visual} with 4447 challenging images, and DUTS-TE \cite{wang2017learning} with 5019 images for testing.


\paragraph{Comparison to state-of-the-art methods}
We compare our method with the 15 most recent state-of-the-art models, including RASv2 \cite{chen2020reverse}, F3Net \cite{wei2019f3net}, ITSD \cite{zhou2020interactive}, GateNet \cite{Zhao2020SuppressAB}, PAGE \cite{wang2019PAGE}, AFNet \cite{feng2019attentive},  PoolNet \cite{liu2019simple}, BASNet \cite{qin2019basnet}, CPD \cite{wu2019cascaded}, EGNet \cite{zhao2019egnet}, PAGRN \cite{zhang2018progressive}, and PiCANet \cite{liu2018picanet}.
For fair comparison, all saliency maps are provided by the authors of these works or computed by their released code with recommended configurations. 

Fig. \ref{fig:VC} shows visual comparisons of the proposed method and other SOTA methods, demonstrating that our method can achieve more accurate results.
The precision--recall curve (PR) in Fig. \ref{fig_rgb_pr} shows that our method outperforms other methods on most datasets while only performing slightly worse than ITSD \cite{zhou2020interactive} on the DUT-OMRON dataset.
Results in Table \ref{tab:rgb} show that the proposed method obtains the best results on all datasets in terms of all metrics.
This fact demonstrates the superiority of our method in the task of RGB image SOD.

\subsection{RGB-D image SOD}
\paragraph{Datasets}
Seven popular datasets are used here: LFSD \cite{li2014saliency} (100 indoor/outdoor images collected by a Lytro IIlum camera), RGBD135 \cite{cheng2014depth} (135 indoor images collected by Microsoft Kinect), STEREO \cite{niu2012leveraging} (797 stereoscopic images obtained from the Internet), SSD \cite{zhu2017three} (80 multi-object images from movies), DUT-RGBD \cite{piao2019depth} (test set of 400 multi-object images), NJUD \cite{ju2014depth} (test set of 500 images) and NLPR \cite{peng2014rgbd} (test set of 300 images collected by Microsoft Kinect).
\paragraph{Comparisons to state-of-the-art methods}
We compared the performance of the proposed method with 12 recent state-of-the-art methods, including DMRA \cite{piao2019depth}, CPFP \cite{zhao2019contrast}, A2dele \cite{Piao2020A2deleAA}, CoNet \cite{ji2020accurate}, JL-DCF \cite{fu2020jl}, PGAR \cite{chen2020progressively}, S2MA \cite{Liu2020LearningSS}, UCNet \cite{Zhang2020UCNetUI}, SSF \cite{zhang2020select}, D3Net \cite{Fan2020RethinkingRS},  ATSA \cite{zhang2020asymmetric}, and ICNet \cite{li2020icnet}.
For fair comparison, all saliency maps are provided by the authors of these works or computed by their released code with recommended configurations.  

Visual comparisons are shown in Fig. \ref{fig:VC}, where the proposed method is compared with some of the SOTA.
This figure demonstrated our model can obtain more accurate results than other methods.
The quantitative results listed in Table \ref{tab:rgb} demonstrate that our method outperforms the other methods on most datasets in terms of most metrics, except that the performance of our method on the RGBD135 dataset is not as good as on other datasets.
We will discuss the reason for this in Sec.\ref{FC} (Failure Cases).

\begin{table}[t]
	\centering
	\caption{RGB-Ablation. CA denotes the cross-attention block, ChA denotes the channel attention block, MS denotes the multi-scale block, and AF denotes the attentive fusion block. `w/o' means `without' this loss function or operation.}
	\resizebox{85mm}{23mm}{
		\begin{tabular}{rrrrrrrrr}
			\toprule
			\multicolumn{1}{|c|}{\multirow{2}[4]{*}{\textbf{Model}}} & \multicolumn{4}{c|}{\textbf{Component}} & \multicolumn{4}{c|}{\textbf{HKI-IS}} \\
			\cmidrule{2-9}    \multicolumn{1}{|c|}{} & \multicolumn{1}{c}{CA} & \multicolumn{1}{c}{ChA} & \multicolumn{1}{c}{MS} & \multicolumn{1}{c|}{AF} & \multicolumn{1}{c}{maxF $ \uparrow $ } & \multicolumn{1}{c}{mF $ \uparrow $ } & \multicolumn{1}{c}{wF $ \uparrow $ } & \multicolumn{1}{c|}{Sm $ \uparrow $ } \\
			\midrule
			\multicolumn{1}{|l|}{baseline} &       &       &       & \multicolumn{1}{c|}{} & 0.916 & 0.874 & 0.867 & \multicolumn{1}{r|}{0.9} \\
			\multicolumn{1}{|l|}{+CA} & \multicolumn{1}{c}{\textbf{$ \checkmark $}} &       &       & \multicolumn{1}{c|}{} & 0.934 & 0.901 & 0.894 & \multicolumn{1}{r|}{0.914} \\
			\multicolumn{1}{|l|}{+CA+ChA} & \multicolumn{1}{c}{\textbf{$ \checkmark $}} & \multicolumn{1}{c}{\textbf{$ \checkmark $}} &       & \multicolumn{1}{c|}{} & 0.936 & 0.904 & 0.899 & \multicolumn{1}{r|}{0.918} \\
			\multicolumn{1}{|l|}{+CA+ChA+MS} & \multicolumn{1}{c}{\textbf{$ \checkmark $}} & \multicolumn{1}{c}{\textbf{$ \checkmark $}} & \multicolumn{1}{c}{\textbf{$ \checkmark $}} & \multicolumn{1}{c|}{} & \textbf{0.939} & 0.911 & 0.904 & \multicolumn{1}{r|}{0.921} \\
			\midrule
			\multicolumn{1}{|c|}{w/o Edge} & \multicolumn{1}{c}{\textbf{$ \checkmark $}} & \multicolumn{1}{c}{\textbf{$ \checkmark $}} & \multicolumn{1}{c}{\textbf{$ \checkmark $}} & \multicolumn{1}{c|}{\textbf{$ \checkmark $}} & 0.938 & 0.914 & 0.903 & \multicolumn{1}{r|}{0.92} \\
			\multicolumn{1}{|c|}{w/o BCE} & \multicolumn{1}{c}{\textbf{$ \checkmark $}} & \multicolumn{1}{c}{\textbf{$ \checkmark $}} & \multicolumn{1}{c}{\textbf{$ \checkmark $}} & \multicolumn{1}{c|}{\textbf{$ \checkmark $}} & 0.937 & 0.917 & \textbf{0.909} & \multicolumn{1}{r|}{0.917} \\
			\multicolumn{1}{|c|}{w/o IoU} & \multicolumn{1}{c}{\textbf{$ \checkmark $}} & \multicolumn{1}{c}{\textbf{$ \checkmark $}} & \multicolumn{1}{c}{\textbf{$ \checkmark $}} & \multicolumn{1}{c|}{\textbf{$ \checkmark $}} & 0.938 & 0.899 & 0.887 & \multicolumn{1}{r|}{0.92} \\
			\multicolumn{1}{|c|}{w/o MLS} & \multicolumn{1}{c}{\textbf{$ \checkmark $}} & \multicolumn{1}{c}{\textbf{$ \checkmark $}} & \multicolumn{1}{c}{\textbf{$ \checkmark $}} & \multicolumn{1}{c|}{\textbf{$ \checkmark $}} & 0.938 & 0.912 & 0.904 & \multicolumn{1}{r|}{0.921} \\
			\midrule
			\multicolumn{1}{|c|}{\textbf{Full}} & \multicolumn{1}{c}{\textbf{$ \checkmark $}} & \multicolumn{1}{c}{\textbf{$ \checkmark $}} & \multicolumn{1}{c}{\textbf{$ \checkmark $}} & \multicolumn{1}{c|}{\textbf{$ \checkmark $}} & \textbf{0.939} & \textbf{0.918} & 0.907 & \multicolumn{1}{r|}{\textbf{0.922}} \\
			\midrule
			&       &       &       &       &       &       &       &  \\
	\end{tabular}}%
	\label{tab:rgb-ablation}%
\end{table}%

\subsection{Ablation Studies}
\begin{figure}
	\begin{center}
		\includegraphics[width=0.45\textwidth]{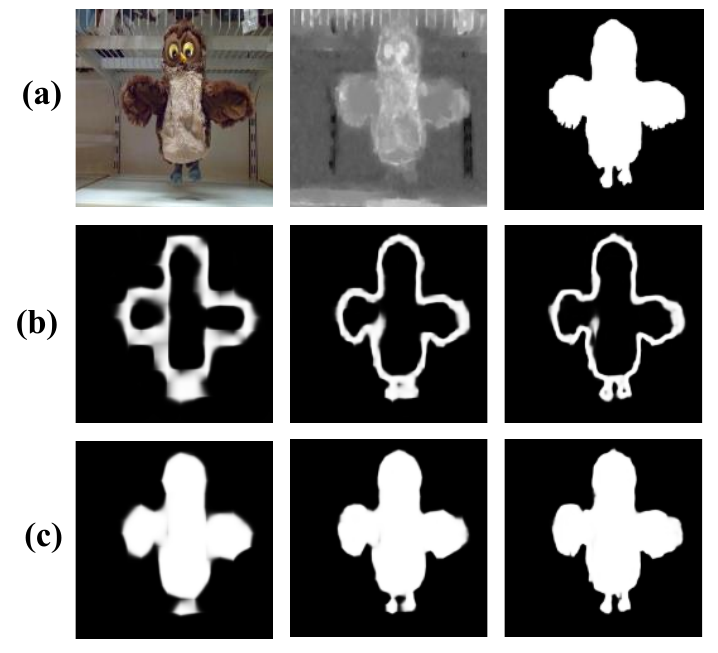}
	\end{center}
	\caption{Context gain after the cross-attention block.  From left to right, (a) shows the images, depth, and ground-truth maps. (b) shows the boundary results ($ E_4, E_3, E_2$ as in Fig.\ref{fig:CFNet}) with the CRACE module computed from the deep to shallow layer. (c) shows the saliency prediction map ($ P_4, P_3, P_2$ as in Fig.\ref{fig:CFNet}) with the CRACE module.}
	\label{fig:CA_out}
\end{figure}

To evaluate the effectiveness of each component, we conducted an additional series of experiments. Table \ref{tab:rgb-ablation} shows the ablation experiment results of our method on RGB image SOD in terms of the maximum, mean, and weighted F-measures, and the structure measure.
As the table shows, the sub-modules of the proposed CRACE module can improve performance.
Moreover, supervision of the boundary, the BCE loss, and the IoU loss also benefit the task, which further validates the proposed structure.
Fig. \ref{fig:CA_out} shows that the results of the saliency outputs and the boundary outputs are gradually better when following the information flow.

Similar conclusions can be drawn from the ablation experiments for the RGB-D image SOD task.
As Table \ref{tab:rgbd-ablation} shows, all sub-modules of the CRACE module facilitated task performances; the edge supervision, and BCE and IoU loss functions also aided performance.
When the depth information is not used, the performance is degraded ($\sim 3\% $) in terms of most metrics.
This fact proves that our CRACE module can appropriately extract and make good use of depth information.

\begin{table}[t]
	\centering
	\caption{RGB-D-Ablation. CA denotes the cross-attention block, ChA denotes the channel attention block, MS denotes the multi-scale block, and AF denotes the attentive fusion block. `w/o' means `without' this loss function or operation.}
	\resizebox{85mm}{23mm}{
		\begin{tabular}{|c|cccc|cccc|}
			\toprule
			\multirow{2}[4]{*}{\textbf{Model}} & \multicolumn{4}{c|}{\textbf{Component}} & \multicolumn{4}{c|}{\textbf{NJUD}} \\
			\cmidrule{2-9}          & CA    & ChA   & MS    & AF    & maxF $\uparrow$ & mF $ \uparrow $  & wF $ \uparrow $    & Sm $ \uparrow $  \\
			\midrule
			\multicolumn{1}{|l|}{baseline} &       &       &       &       & 0.91  & 0.889 & 0.875 & 0.899 \\
			\multicolumn{1}{|l|}{+CA} & \textbf{$ \checkmark $} &       &       &       & 0.915 & 0.902 & 0.886 & 0.904 \\
			\multicolumn{1}{|l|}{+CA+ChA} & \textbf{$ \checkmark $} & \textbf{$ \checkmark $} &       &       & 0.917 & 0.899 & 0.886 & 0.905 \\
			\multicolumn{1}{|l|}{+CA+ChA+MS} & \textbf{$ \checkmark $} & \textbf{$ \checkmark $} & \textbf{$ \checkmark $} &       & 0.919 & 0.903 & 0.891 & 0.907 \\
			\midrule
			w/o Depth & \textbf{$ \checkmark $} & \textbf{$ \checkmark $} & \textbf{$ \checkmark $} & \textbf{$ \checkmark $} & 0.901 & 0.869 & 0.863 & 0.889 \\
			w/o Edge & \textbf{$ \checkmark $} & \textbf{$ \checkmark $} & \textbf{$ \checkmark $} & \textbf{$ \checkmark $} & 0.922 & 0.903 & \textbf{0.892} & 0.907 \\
			w/o BCE & \textbf{$ \checkmark $} & \textbf{$ \checkmark $} & \textbf{$ \checkmark $} & \textbf{$ \checkmark $} & 0.92  & 0.903 & 0.891 & 0.905 \\
			w/o IoU & \textbf{$ \checkmark $} & \textbf{$ \checkmark $} & \textbf{$ \checkmark $} & \textbf{$ \checkmark $} & 0.92  & 0.879 & 0.883 & \textbf{0.912} \\
			w/o MLS & \textbf{$ \checkmark $} & \textbf{$ \checkmark $} & \textbf{$ \checkmark $} & \textbf{$ \checkmark $} &  \textbf{0.923}   & 0.904 & 0.890 & 0.906 \\
			\midrule
			\textbf{Full} & \textbf{$ \checkmark $} & \textbf{$ \checkmark $} & \textbf{$ \checkmark $} & \textbf{$ \checkmark $} & \textbf{0.923} & \textbf{0.905} & \textbf{0.892} & 0.907 \\
			\bottomrule
	\end{tabular}}%
	\label{tab:rgbd-ablation}%
\end{table}%

\subsection{Time Efficiency}
Time efficiency is an important indicator in SOD.
Since the proposed method has a simple structure with only three CRACE modules, our method can efficiently accomplish both tasks.
Table \ref{tab:FPS} compares our method and other representative methods in terms of frame per second (fps).
As the table shows, our method is fast in the inference stage when processing images in real-time ($ \geq25fps$).
Other leading methods can also operate in real time in the inference stage, but they have relatively poor accuracy.

Similar observations are also found for the task of RGB-D SOD.
Our method with Resnet50 can run as fast as 23 fps (near real time), which is as fast as SSF \cite{zhang2020select} and faster than PGAR \cite{chen2020progressively} and S2MA \cite{Liu2020LearningSS}, which all have vgg16 as the backbone.
Overall, considering the performances shown in Tables \ref{tab:rgb} and \ref{tab:rgbd}, our method has promising performance both in efficiency and accuracy.

\subsection{Failure Cases}
\label{FC}
\begin{figure}
	\begin{center}
		\includegraphics[width=0.45\textwidth]{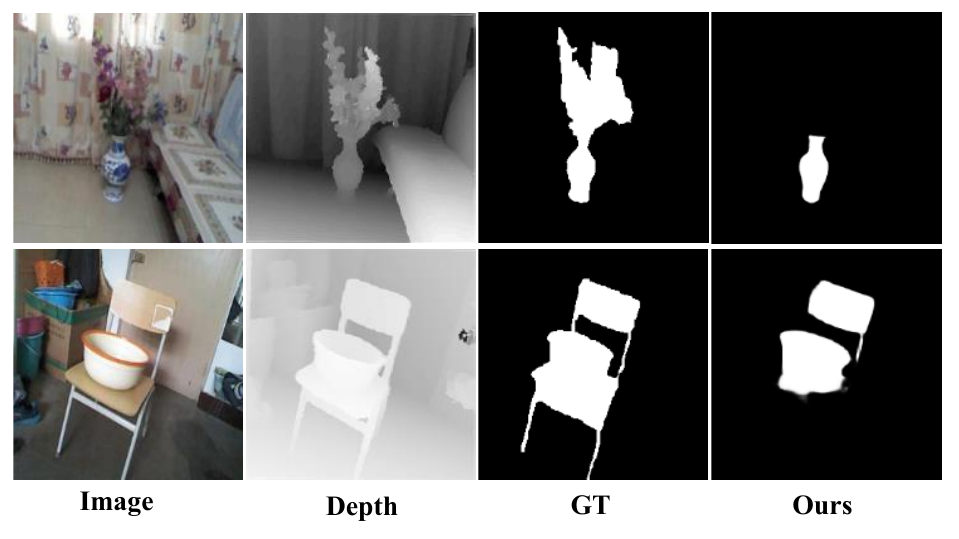}
	\end{center}
	\caption{Failure cases on two images selected from the RGBD135 dataset.}
	\label{fig:FailureCases}
\end{figure}

Fig. \ref{fig:FailureCases} shows two failure cases on the RGBD135 dataset since our model seems weaker than others on this dataset.
The top row shows a complex example where the depth map highlights the flowers and vase, while only the vase is considered as a salient object by our model.
The bottom row shows that our method cannot segment the chair as accurate as the ground-truth, but the washbasin is successfully found.
This is probably because our method pays more attention to the semantics that RGB information provided in these scenes, where the vase and washbasin are segmented. 
Unfortunately, there are 22 images that contain both the vase and washbasin in this dataset (for a total of 135 images), which significantly degrades the performance of our model.
In future works, we will try to improve the model by adaptively balance the contributions of RGB and depth information so that it can achieve good performance on such images.

\section{Conclusion}
In this work, we first proposed a new CRACE module. Based on this module, we proposed a simple unified but effective network to accomplish SOD for both RGB and RGB-D images.  
The proposed CRACE module makes good use of global information from higher layers and local information from lower layers by extracting and applying a cross-attention mechanism. Cross-modal features are also able to be processed within this module.
Experimental results show that the proposed unified structure achieves the best overall performance compared with existing state-of-the-art methods on both RGB and RGB-D image SOD tasks.

\begin{table}[t]
	\centering
	\caption{Time Efficiency.}
	\resizebox{85mm}{13mm}{
		\begin{tabular}{|l|c|cccc|}
			\toprule
			\textbf{RGB-Methods} & \textbf{Ours} & \textbf{RASv2} & \textbf{EGNet} & \textbf{PoolNet} & \textbf{F3Net} \\
			\midrule
			Backbone & ResNet50 & ResNet50 & ResNet50 & ResNet50 & ResNet50 \\
			Speed (fps) & 37    & 40   & 8     & 25    & 34 \\
			\midrule
			\textbf{RGBD-Methods} & \textbf{Ours} & \textbf{SSF} & \textbf{D3Net} & \textbf{PGAR} & \textbf{S2MA} \\
			\midrule
			Backbone & ResNet50 &VGG16 & VGG16 & VGG16 & VGG16 \\
			Speed (fps) & 23    & 23    & 40    & 18    & 16 \\
			\bottomrule
	\end{tabular}}%
	\label{tab:FPS}%
\end{table}%

\section*{Acknowledgment}
This work was supported by Guangdong Key R\&D Project (\#2018B030338001) and Natural Science Foundations of China (\#61806041, \#61703075).
We also thank the LetPub for its linguistic assistance during the preparation of this manuscript.

\ifCLASSOPTIONcaptionsoff
  \newpage
\fi



\bibliographystyle{IEEEtran}
\bibliography{ppbib}{}
%

%

%

%
\begin{IEEEbiography}[{\includegraphics[width=1in,height=1.25in,clip,keepaspectratio]{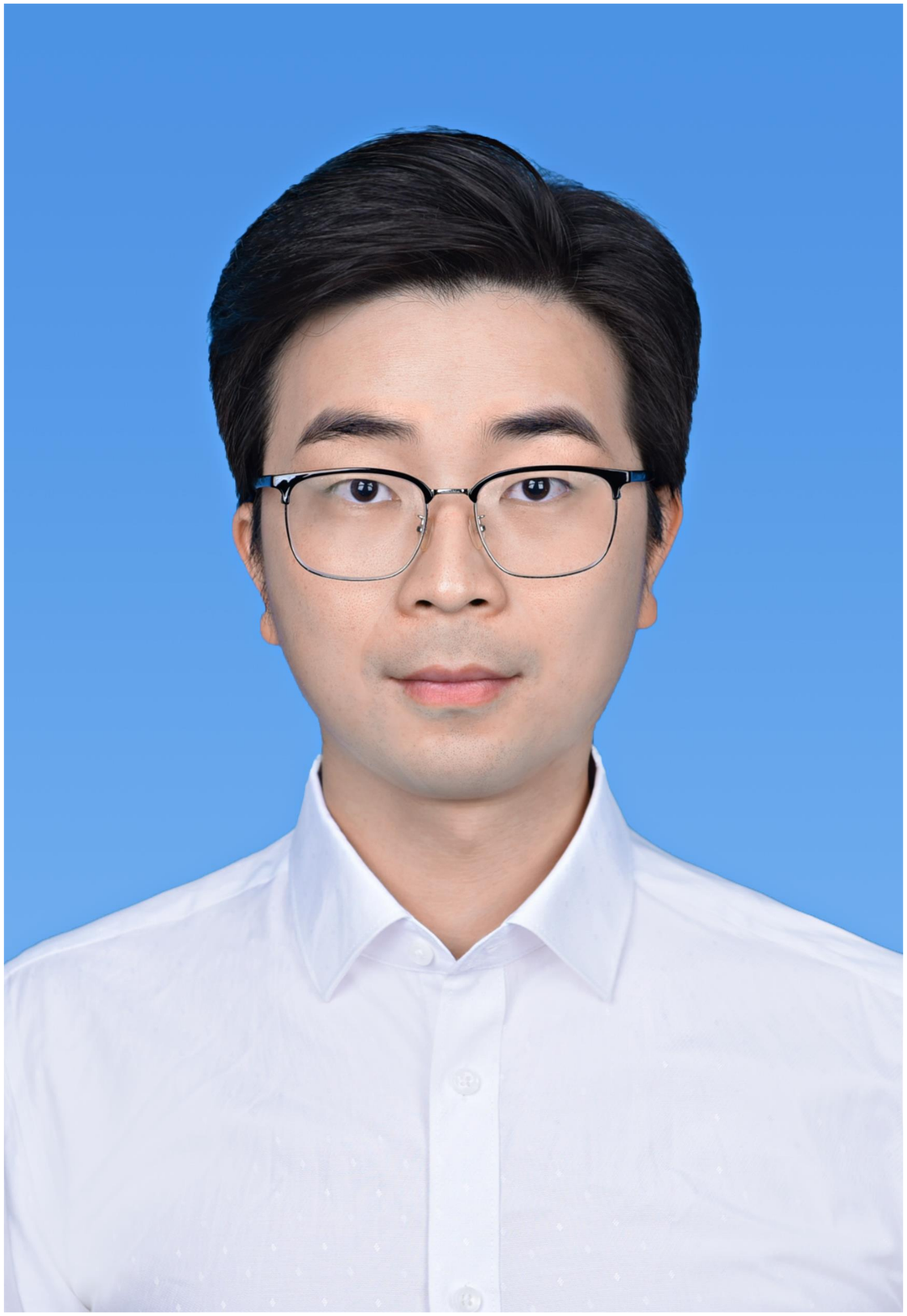}}]{Peng Peng}
	received the BS degree in biomedical engineering from the University of Electronic Science and Technology of China (UESTC), in 2015.
	He is now pursuing his Ph.D. degree in UESTC.
	His research interests include the visual perceptual organization in early vision, brain-inspired computer vision, saliency detection, image segmentation and machine learning.
\end{IEEEbiography}



\begin{IEEEbiography}[{\includegraphics[width=1in,height=1.25in,clip,keepaspectratio]{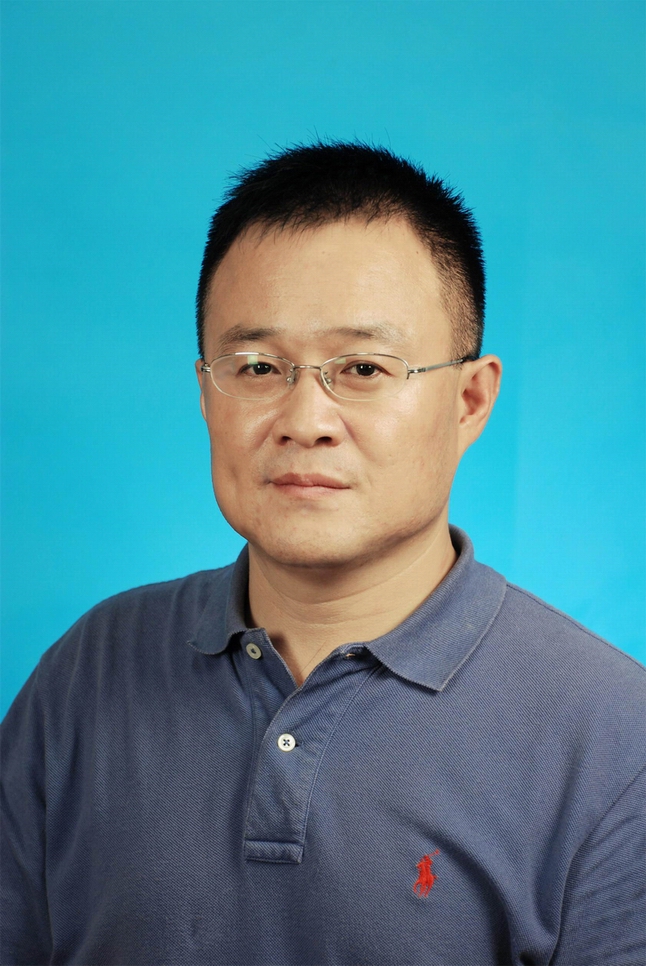}}]{Yong-Jie Li}
	(M'14-SM'17) received the Ph.D. degree in biomedical engineering from the University of Electronic  Science and Technology of China (UESTC) in 2004.
	He is currently a Professor with the MOE Key Laboratory for Neuroinformation, School of Life Science and Technology, UESTC, China. His research interests include visual mechanism modeling, and the applications in image processing and computer vision. 
\end{IEEEbiography}





\end{document}